\documentclass{article}

\usepackage[T1]{fontenc}
\usepackage{graphicx}
\usepackage[small]{caption}
\usepackage{color}
\usepackage{setspace}
\usepackage{cite} 

\begin{document}

\doublespacing

\title{Crossings as a side effect of dependency lengths}

\author{Ramon Ferrer-i-Cancho$^{1,*}$ \& Carlos G\'omez-Rodr\'iguez$^2$ \\
~\\
\begin{minipage}[c]{\textwidth}
{\small $^1$ Complexity \& Qualitative Linguistics Lab, LARCA Research Group,
Departament de Ci\`encies de la Computaci\'o, Universitat Polit\`ecnica de Catalunya, Campus Nord, Edifici Omega. Jordi Girona Salgado 1-3. 08034 Barcelona, Catalonia, Spain \\ 
$^2$ LyS Research Group, Departamento de Computaci\'on, Facultade de Inform\'atica, Universidade da Coru\~na, Campus de A Coru\~na, 15071 A Coru\~na, Spain\\
~\\
$*$ To whom correspondence should be addressed. E-mail: rferrericancho@cs.upc.edu. Phone: +34 934134028}
\end{minipage}
}

\date{}

\maketitle

\begin{abstract}
The syntactic structure of sentences exhibits a striking regularity: dependencies tend to not cross when drawn above the sentence. We investigate two competing explanations. 
The traditional hypothesis is that this trend arises from an independent principle of syntax that reduces crossings practically to zero. 
An alternative to this view is the hypothesis that crossings are a side effect of dependency lengths, i.e. sentences with shorter dependency lengths should tend to have fewer crossings.
We are able to reject the traditional view in the majority of languages considered. The alternative hypothesis can lead to a more parsimonious theory of language.
\end{abstract}


{\bf Nontechnical, jargon-free summary: }
Syntactic relations between words (e.g., the one that links a verb with its subject) exhibit a strong tendency to not cross when drawn as arrows above the sentence. Traditionally, this has been assumed to result from an independent principle of syntax that reduces crossings practically to zero. An alternative view is that the trend arises naturally from the preference in human languages for word orders that keep related words close together. Our statistical analysis discards the traditional view in the majority of languages considered. The alternative approach can lead to a simpler theory of language.

{\bf 26 pages, 2 figures and 3 tables. }

\section{Introduction}

\label{introduction_section}

\begin{figure}
\begin{center}
\includegraphics[scale = 1]{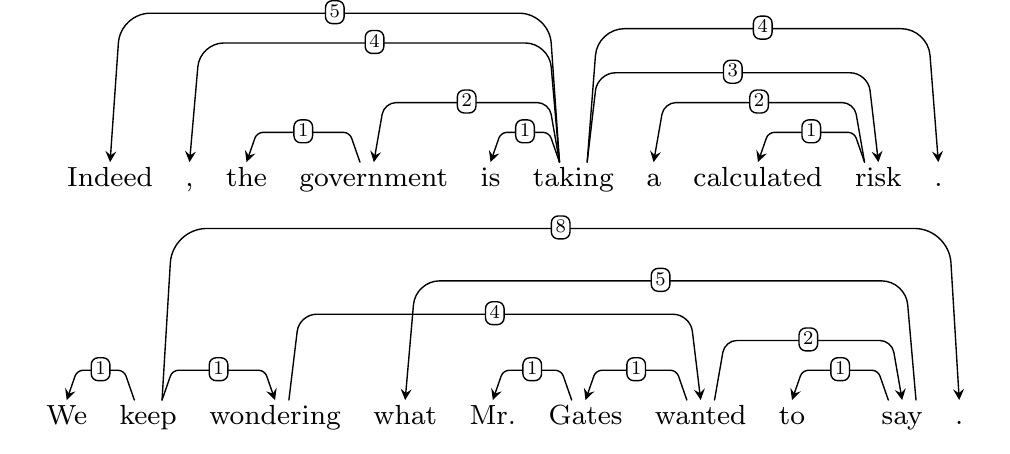}
\end{center}
\caption{\label{non_crossing_dependencies_figure} 
Two sentences with Stanford annotations from HamleDT 2.0 \protect\cite{HamledTStanford}. Dependencies are labelled with their length (in tokens). For the sentence on top, the sum of dependency lengths is $D = 23$ and the number of crossings is $C = 0$; $D = 24$ and $C = 1$ for the sentence at the bottom.}
\end{figure}

One of the main goals of complexity science is to provide parsimonious explanations for statistical patterns that are observed in nature \cite{Christensen2005a,Newman2010a}. Here we pay attention to 
a striking regularity of the syntactic structure of sentences that was reported in the 1960s:
dependencies tend to not cross when drawn above the sentence \cite{Lecerf1960a,Hays1964}, as shown in Fig. \ref{non_crossing_dependencies_figure}. The absence of crossings is known as planarity, a feature that is intimately related with another property of syntactic dependency trees: projectivity \cite{KuhNiv06}. Projectivity is a particular case of planarity where no dependency covers the root. Interestingly, real sentences that are planar tend to be projective \cite{KuhNiv06,Havelka2007,GomNivCL2013}.
Here we investigate two competing hypotheses for the origins of non-crossing dependencies. 

The traditional hypothesis is that the low frequency of dependency crossings arises from an independent principle of syntax that reduces crossings practically to zero. This view is held by theories of grammar where crossings are not allowed \cite{sleator93,hudson07,Tanaka97,KyotoCorpus} and also by parsing frameworks where non-crossing dependencies are not allowed or subject to hard constraints \cite{sleator93,Nivre2003a,Carreras2007a,Zhang2011a,Chen2014a,Dyer2015a}. 
It is also shared by research on dependency length minimization where annotations with crossings are discarded \cite{Gildea2010a} 
and actual dependency lengths are compared with two kinds of baselines where crossings are not allowed or are subject to hard constraints: random orderings and optimal dependency lengths \cite{Temperley2008a,Liu2008a,Park2009a,Gildea2010a,Futrell2015a,Gulordava2015}. 
The traditional view is convenient for simplicity and computational reasons: efficient algorithms for non-crossing dependencies or limited violations are available \cite{Hochberg2003a,Gildea2007a,Park2009a} and is justified by the low frequency of crossings in real languages \cite{Temperley2008a,Gildea2010a}. 

An alternative to this view is the hypothesis that crossings are a side effect of dependency lengths \cite{Ferrer2014c,Ferrer2014f}. This hypothesis predicts that dependencies should tend to not cross, combining a tendency for shorter dependency lengths to have fewer crossings and the fact that dependencies are actually short.
This challenges the dogma that unconstrained dependency length minimization ``does not take into account constraints of projectivity or mild context-sensitivity'' \cite{Park2009a}; and is coherent with the trends towards diachronic reduction of the proportion of crossings in conjunction with dependency length minimization that have been observed on English \cite{Tily2010a} and also recently on Latin and Ancient Greek \cite{Gulordava2015}.
 
Here we will evaluate these two hypotheses making emphasis on the validity of the traditional view. We will formalize the traditional view as a null hypothesis and the alternative view as an alternative hypothesis. 
With the help of a collection of dependency treebanks of thirty different languages, we will 
show that the null hypothesis of the traditional view is rejected for a large majority of treebanks.

\section{Formalization of the problem}

\label{modelling_section}

Suppose that $C$ is the number of crossings of a sentence and that $n$ is its number of words. We define $E_{TB}[C|n, D]$ as the expectation of $C$ conditioning on sentences of a treebank (TB) that have length $n$ and their sum of dependency lengths is $D$. Then the traditional view can be recast simply as
\begin{equation}
E_{TB}[C | n, D] = a_{TB}(n),
\label{ban_of_crossings_model_equation}
\end{equation}   
where $a_{TB}$ is a constant with respect to $D$ that depends on $n$. For the particular case of a complete ban on crossings, $a_{TB}(n) = 0$ for all $n$. The Appendix provides a derivation of Eq. \ref{ban_of_crossings_model_equation}, including a detailed explanation of why $a_{TB}$ depends on $n$ in general. Notice that $a_{TB}(n)$ is constant for all trees of length $n$ and bear in mind that we will test Eq. \ref{ban_of_crossings_model_equation} on sentences of the same length.

The fact that $a_{TB}(n) = E_{TB}[C | n]$ \cite{Ferrer2012h} allows one to formulate the traditional view equivalently as 
\begin{equation}
E_{TB}[C| n, D] = E_{TB}[C | n].
\label{mean_independence_equation}
\end{equation}
Thus, given a treebank and a sentence length $n$, the traditional hypothesis predicts that a sentence will have, on average, a number of crossings that coincides with the mean number of crossings of the sentences of length $n$. Accordingly, the alternative view is modeled by
\begin{equation}
E_{TB}[C|n, D ] = g_{TB}(n, D),
\label{simple_alternative_model_equation}
\end{equation}
where $g_{TB}(n, D)$ is a strictly monotonically increasing function of $D$ when $n$ remains constant. 
In this article, we want to remain agnostic about the exact mathematical form of $g_{TB}(n, D)$. Our focus is on the validity of the traditional view. Concerning the alternative view, we are only interested in the sign of the correlation between $C$ and $D$. A positive correlation provides support for the hypothesis that crossings are a side effect of dependency lengths. 
Note that a positive correlation between $D$ and $C$ has been shown empirically in real syntactic dependency trees, but assuming unrealistic word orders (in particular, uniformly random linear arrangements) \cite{Ferrer2014f}. This correlation has been supported using theoretical arguments that show that reducing the length of a dependency is likely to imply a reduction in the probability that two edges cross, assuming random arrangements that are also unrealistic \cite{Ferrer2014f,Ferrer2014c}. 
The limitations of previous research on the hypothesis raise the question of whether such a correlation still holds when considering linear arrangements that are actually reached. For the first time, here we will investigate the correlation between $D$ and $C$ involving their joint distribution in real linear arrangements of syntactic dependency trees. Put differently, here we are testing a new condition that is vital to evaluate the hypothesis that $C$ is a side effect of dependency lengths, and not a consequence of an autonomous principle of syntax that disallows or bounds crossings.  
  
Eq. \ref{mean_independence_equation} is interesting because it indicates that 
the traditional view is equivalent to $C$ being mean independent of $D$ when $n$ is given, in the language of probability theory \cite[p. 67]{Poirier1995a}. From the perspective of statistical hypothesis testing, the traditional view is a null hypothesis (mean independence), while the alternative view (a positive correlation between $D$ and $C$) is an alternative hypothesis. 

Although the autonomous bound on crossings has never been explicitly formulated as in Eq. \ref{ban_of_crossings_model_equation} or \ref{mean_independence_equation}, a mathematical definition that can be used for testing following standard statistical methods is not forthcoming. In the game of science, hypotheses must be precise enough to be falsified \cite{Popper1963a}. 
One could argue that Eq. \ref{ban_of_crossings_model_equation} or \ref{mean_independence_equation} are a particular interpretation of an autonomous bound on crossings, perhaps a very narrow one. However, is easy to show that a ban on crossings, i.e. $a_{TB}(n)=0$, and 
Eq. \ref{mean_independence_equation} with $E_{TB}[C | n, D] = 0$ are equivalent once one focuses on sentences of the same length: 
\begin{itemize}
\item
If $E_{TB}[C | n, D] = 0$ then $C = 0$ for any tree of $n$ vertices because $C \geq 0$ by definition. 
\item
If $C = 0$ for any tree of $n$ vertices, then $E_{TB}[C | n, D] = 0$ obviously.
\end{itemize}    
The null hypothesis with $a_{TB}(n) \ge 0$ (Eq. \ref{ban_of_crossings_model_equation}) is simply a relaxation of the ban.  

\begin{figure}
\begin{center}
\includegraphics[scale = 0.6]{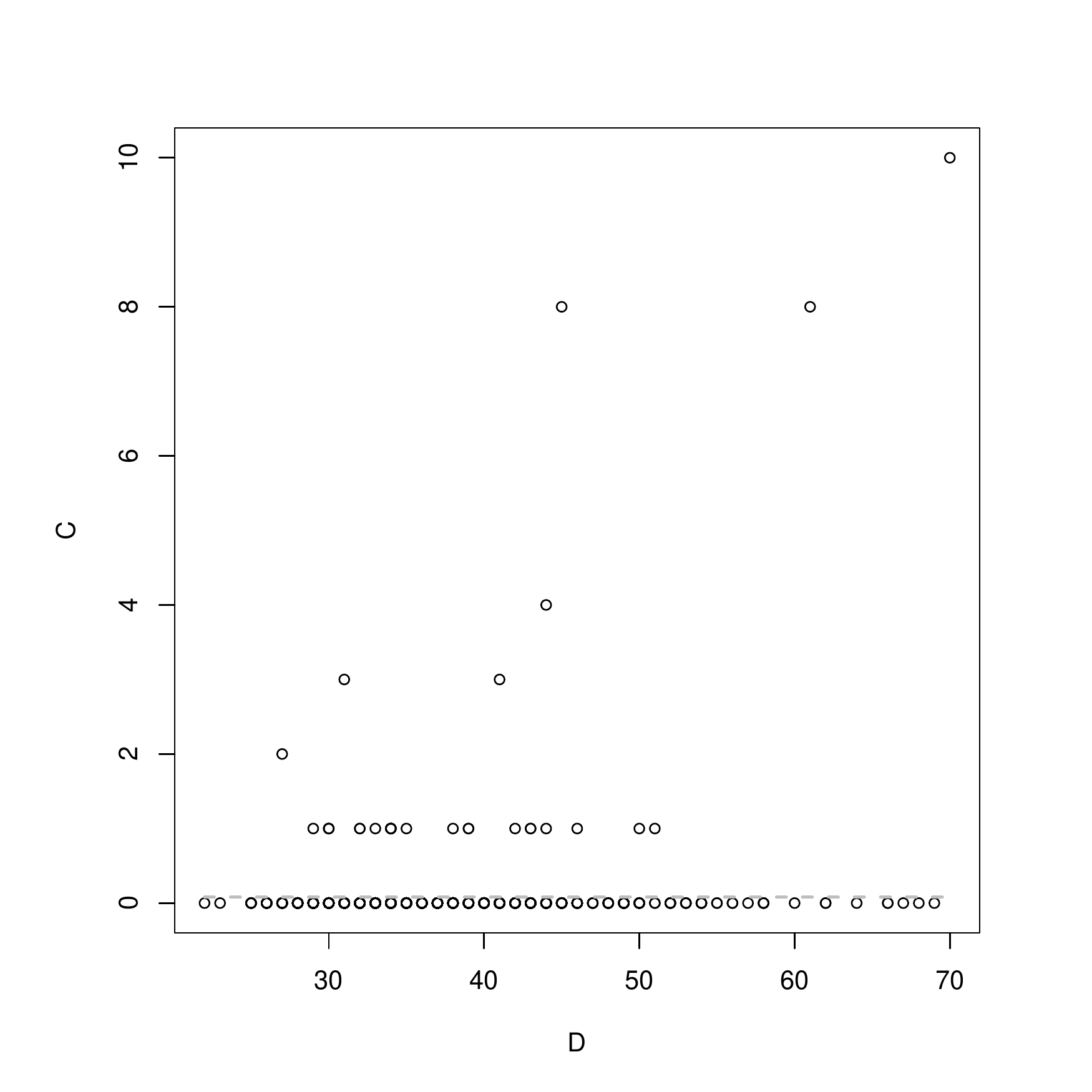}
\end{center}
\caption{\label{lengths_vs_crossings_figure} Crossings ($C$) versus sum of dependency lengths ($D$) in sentences of length 18 in an English treebank (we use Prague dependencies from HamleDT 2.0, see Section \ref{materials_section}). 18 is the typical sentence length in this treebank. The average prediction made by the null hypothesis is also shown (gray dashed line). }
\end{figure}

Fig. \ref{lengths_vs_crossings_figure} compares the relationship between $D$ and $C$ in sentences of length 18 in an English dependency treebank. In this case, the traditional view is 
\begin{equation}
E_{TB}[C|n, D ] = a_{TB}(n)
\end{equation}
with $a_{TB}(n) = 0.08$, the mean number of crossings in sentences of length 18 in that treebank. This very low number casts doubts on the adequacy of the null hypothesis for the large values of $C$ that are found especially for large values of $D$ in Fig. \ref{lengths_vs_crossings_figure}. 
The Kendall $\tau$ correlation between $C$ and $D$ is $\tau = 0.03$ (p-value = $0.28$) indicating a weak but positive tendency of $C$ to increase as $D$ increases. In this article, we will study collections of sentences with syntactic dependency annotations (treebanks) of different languages, to check if the number of positive 
$\tau$ correlations across sentence lengths is significantly high. If that happens, we will conclude that an autonomous bound on crossings (Eq. \ref{ban_of_crossings_model_equation}) does not hold in general for that treebank. 

It is tempting to think that Eq. \ref{mean_independence_equation} is impossible to satisfy and thus the rejection of the null hypothesis is inevitable. However, notice three facts. First, $E[C | n, D] = 0$ can be satisfied at least for the particular case that the trees are star trees: in that case $C = 0$ while \cite{Ferrer2013e} 
\begin{equation}
\frac{n^2 - n \mbox{~mod~} 2}{4} \leq D \leq \frac{n(n-1)}{2}.
\end{equation} 

Second, for any given treebank, the null hypothesis is also satisfied by any reordering of the words in the sentences that enforces $C=0$. Concrete examples come from Hochberg \& Stallmann's algorithm, that provides minimum linear arrangements without crossings \cite{Hochberg2003a} as well as 
the random and optimal projective linearization algorithms employed in the dependency length research reviewed in Section \ref{introduction_section} (e.g., \cite{Gildea2010a,Futrell2015a}). 

Third, our analysis will show that the null hypothesis could not be rejected in all treebanks (some preliminary evidence is provided by Fig. \ref{lengths_vs_crossings_figure}, that shows a correlation between $D$ and $C$ that is not statistically significant).  

We would like to emphasize that the goal of this article is not to predict the actual number of crossings with great accuracy as in related work \cite{Ferrer2014c,Gomez2016a}
but to examine the validity of 
the customary assumption of an autonomous bound on crossings 
with a simple (and statistically sound) approach. 
$D$ is a rough predictor of crossings because the probability that two dependencies cross is determined by their individual lengths and whether they share vertices or not \cite{Ferrer2014c, Ferrer2014f}. $D$ can be seen as a lossy compression of the dependency lengths of a sentence into a single value. Furthermore, other factors such as chunking can have an important role in the formation of crossings \cite{Lu2016a}.
Thus, it is rather surprising that the rough predictions that $D$ offers allow us to reject the traditional view in the majority of treebanks, as we will see.

\section{Materials and methods}

\subsection{Materials}

\label{materials_section}

We employ HamleDT 2.0, a collection of dependency treebanks of 30 different languages \cite{HamledTStanford}.
The collection provides sentences with syntactic dependency annotations following two different criteria: Prague dependencies \cite{PDT20} and Stanford dependencies \cite{UniversalStanford}. This collection allows one to explore a set of typologically diverse languages and control for the effect of annotation criteria.

Each syntactic dependency structure in the treebanks was preprocessed by removing nodes corresponding to punctuation tokens. To preserve the syntactic structure of the rest of the nodes, non-punctuation nodes that had a punctuation node as their head were attached as dependents of their nearest non-punctuation ancestor. Null elements, which appear in the Bengali, Hindi and Telugu corpora, were also subject to the same treatment as punctuation.

After this preprocessing, syntactic dependency structures that did not define a tree were removed.   
The reason is that we wanted to avoid the statistical problem of mixing trees with other kinds of graphs, e.g. the potential number of crossings depends on the number of edges \cite{Ferrer2013b,Ferrer2013d,Ferrer2014f}.

\subsection{Methods}

For each sentence length of a treebank, we want to investigate if the null hypothesis that $C$ is mean independent of $D$ actually holds. This can be tested with the help of the Kendall $\tau$ correlation between $C$ and $D$. Suppose that $c_1$ and $c_2$ are two observations of $C$ and $d_1$ and $d_2$ are two observations of $D$. Then $(c_1,d_1)$ and $(c_2,d_2)$ are said to be concordant if $(c_1-c_2)(d_1-d_2)>0$ (the ranks of both elements agree), and discordant if $(c_1-c_2)(d_1-d_2)<0$ (they disagree). Then, Kendall $\tau$ correlation is defined as \cite{Conover1999a} 
\begin{equation}
\tau = \frac{N_c - N_d}{N_0},
\label{Kendall_tau_correlation_equation}
\end{equation}
where $N_c$ is the number of concordant pairs, $N_d$ is the number of discordant pairs and $N_0$ is the total number of pairs. 

For each treebank, we calculated the Kendall $\tau$ correlation between $D$ and $C$ for every sentence length $n$. Sentence lengths that met at least one of the following conditions were excluded from the analysis: 
\begin{itemize}
\item
$n < 4$, because $C = 0$ for them \cite{Ferrer2013b}.
\item
Lengths that were represented by less than two sentences, because $N_0=0$ and then $\tau$ is not properly defined.
\end{itemize}
Then we calculated 
$p(\tau \geq 0)$, the proportion of sentence lengths where $\tau \geq 0$. 
If $p(\tau \geq 0)$ is sufficiently high then the null hypothesis of mean independence is rejected.
The significance of $p(\tau \geq 0)$ was determined with the help of a Monte Carlo method that takes as input the vectors $\vec{D^n} = \{d_1^n, \dots ,d_i^n, \dots ,d_m^n\}$ and $\vec{C^n} = \{c_1^n, \dots ,c_i^n, \dots ,c_m^n\}$ of every sentence length $n$ ($d_i^n$ and $c_i^n$ are, respectively, the sum of dependency lengths and number of crossings of the $i$-th sentence of length $n$). This method consists of generating $T$ randomizations of the input vectors and estimating the p-value of the test as the proportion of times that $p_c(\tau \geq 0) \geq p(\tau \geq 0)$, where $p_c(\tau \geq 0)$ is the value of $p(\tau \geq 0)$, over $T$ randomizations of the vectors. A randomization consists of replacing the vector $\vec{D^n}$ for each sentence length with a uniformly random permutation.
For this article, we use $T = 10^4$ and a significance level of 0.05.
     
We could have determined the significance of $p(\tau \geq 0)$ by means of a binomial test: 
under the assumption of independence between $D$ and $C$ and assuming that there are no ties among values, the probability that $\tau \geq 0$ is $1/2$ \cite{Prokhorov2001a}. However, ties of $C$ abound (many sentences have $C = 0$, see also Table \ref{sentence_lengths_without_crossings_table}). For this reason, the Monte Carlo test above yields a more accurate estimation of the true p-value. 
 
It is convenient to split $p(\tau \geq 0)$ as $p(\tau > 0) + p(\tau = 0)$ and inspect $p(\tau = 0)$ because Kendall $\tau = 0$ is due to $N_c = N_d$ (recall Eq. \ref{Kendall_tau_correlation_equation}). High p-values of $p(\tau \geq 0)$ could be due to high $p(\tau = 0)$, which in turn would be due to $C=0$ for many sentence lengths. To see it, consider the following extreme case: 
a treebank where $C=0$ in all sentences. In that case, $N_c = N_d = 0$ for all sentence lengths and then   
$p(\tau \geq 0) = p(\tau = 0)$. Interestingly, $\tau$ would remain zero for all sentence lengths after randomization and then the p-value of the Monte Carlo test would be 1. That has been the case of the Romanian treebank with Prague dependencies (Tables \ref{dependencies_table} and \ref{sentence_lengths_without_crossings_table}).

\section{Results}

Table \ref{dependencies_table} shows that 
$p(\tau \geq 0)$ is significantly high in about three fourths of the languages for Prague dependencies (eight treebanks have a p-value above the significance level) and to a much larger extent for the Stanford dependencies (only five treebanks have a p-value above the significance level). Thus, there is a minority of languages where there is not enough support for the hypothesis that crossing dependencies are a side effect of dependency lengths. Interestingly, $p(\tau = 0)$ is especially high in the treebanks where $p(\tau \geq 0)$ is not significantly high. 
A possible explanation for the failure of the alternative view in those treebanks is that $C=0$ in the majority of sentence lengths. 
Let us call $p_0$ the proportion of sentence lengths where all sentences have $C=0$. 
Table \ref{sentence_lengths_without_crossings_table} indicates that the five treebanks where $p(\tau \geq 0)$ is not significantly high for Stanford dependencies coincide with the five treebanks with the largest $p_0$. The situation for Prague dependencies is similar: the top six largest values of $p_0$ are taken by six treebanks where $p(\tau \geq 0)$ is not significantly high. In the treebanks where $p(\tau \geq 0)$ is not significantly high we have $p(\tau = 0) = p_0$ in practically all cases, although $p(\tau = 0) \geq p_0$ {\em a priori}.
Indeed, the average $p_0$ is significantly high in the subset of the treebanks where the null hypothesis could not be rejected (Table \ref{meta_analysis_table}).

\begin{table}
\begin{center}
{\footnotesize
\begin{tabular}{lllllllll}
 & \multicolumn{4}{c}{Prague} & \multicolumn{4}{c}{Stanford} \\ \cline{2-5}\cline{6-9}
Treebank & $M$        & $p(\tau = 0)$ & $p(\tau > 0)$ & p-value & $M$        & $p(\tau = 0)$ & $p(\tau > 0)$ & p-value \\
\hline
Arabic & 90 & 0.38 & 0.34 & 0.239 & 90 & 0.067 & 0.7 & 0.0001\\
Basque & 33 & 0.091 & 0.85 & $<10^{-4}$ & 33 & 0.03 & 0.82 & 0.0001\\
Bengali & 16 & 0.19 & 0.38 & 0.7372 & 17 & 0.29 & 0.53 & 0.0871\\
Bulgarian & 51 & 0.078 & 0.69 & 0.0006 & 52 & 0.077 & 0.77 & $<10^{-4}$\\
Catalan & 86 & 0.16 & 0.76 & $<10^{-4}$ & 86 & 0.047 & 0.72 & $<10^{-4}$\\
Czech & 73 & 0.027 & 0.78 & $<10^{-4}$ & 74 & 0.054 & 0.76 & $<10^{-4}$\\
Danish & 56 & 0.11 & 0.62 & 0.005 & 57 & 0.07 & 0.77 & $<10^{-4}$\\
Dutch & 52 & 0.019 & 0.81 & $<10^{-4}$ & 52 & 0 & 0.85 & $<10^{-4}$\\
English & 66 & 0.11 & 0.64 & 0.0001 & 66 & 0.045 & 0.64 & 0.0089\\
Estonian & 22 & 0.82 & 0.14 & 0.3027 & 22 & 0.45 & 0.14 & 0.9877\\
Finnish & 33 & 0.15 & 0.79 & $<10^{-4}$ & 33 & 0.091 & 0.88 & $<10^{-4}$\\
German & 72 & 0.042 & 0.71 & $<10^{-4}$ & 72 & 0.014 & 0.64 & 0.0151\\
Greek(ancient) & 53 & 0 & 0.94 & $<10^{-4}$ & 53 & 0 & 0.89 & $<10^{-4}$\\
Greek(modern) & 63 & 0.24 & 0.49 & 0.0358 & 64 & 0.14 & 0.66 & 0.0001\\
Hindi & 58 & 0.069 & 0.78 & $<10^{-4}$ & 58 & 0.086 & 0.74 & $<10^{-4}$\\
Hungarian & 62 & 0.032 & 0.74 & $<10^{-4}$ & 61 & 0.049 & 0.64 & 0.0048\\
Italian & 59 & 0.34 & 0.53 & 0.0002 & 59 & 0.12 & 0.73 & $<10^{-4}$\\
Japanese & 37 & 0.97 & 0 & 1 & 37 & 0 & 0.95 & $<10^{-4}$\\
Latin & 46 & 0 & 0.72 & 0.0042 & 46 & 0.043 & 0.8 & $<10^{-4}$\\
Persian & 71 & 0.028 & 0.25 & 0.9999 & 72 & 0.042 & 0.76 & $<10^{-4}$\\
Portuguese & 79 & 0.063 & 0.71 & $<10^{-4}$ & 79 & 0.089 & 0.75 & $<10^{-4}$\\
Romanian & 38 & 1 & 0 & 1 & 38 & 0.21 & 0.5 & 0.1266\\
Russian & 66 & 0.076 & 0.76 & $<10^{-4}$ & 65 & 0.031 & 0.83 & $<10^{-4}$\\
Slovak & 74 & 0.068 & 0.64 & 0.0008 & 76 & 0.026 & 0.75 & $<10^{-4}$\\
Slovene & 46 & 0.087 & 0.59 & 0.027 & 49 & 0.041 & 0.73 & 0.0001\\
Spanish & 81 & 0.16 & 0.79 & $<10^{-4}$ & 80 & 0.037 & 0.84 & $<10^{-4}$\\
Swedish & 59 & 0.1 & 0.81 & $<10^{-4}$ & 61 & 0.033 & 0.75 & $<10^{-4}$\\
Tamil & 31 & 0.84 & 0.13 & 0.2015 & 31 & 0.68 & 0.26 & 0.0514\\
Telugu & 8 & 0.88 & 0.12 & 0.5315 & 8 & 0.62 & 0.38 & 0.1727\\
Turkish & 43 & 0.07 & 0.67 & 0.0033 & 44 & 0.11 & 0.75 & $<10^{-4}$\\

\end{tabular}
}
\end{center}
\caption{\label{dependencies_table} 
Summary of the analysis of the correlation between $D$ and $C$. For every treebank, we show the number of different sentence lengths considered ($M$), the proportion of sentence lengths where Kendall $\tau$ is equal or greater than zero ($p(\tau = 0)$ and $p(\tau > 0)$, respectively), and the p-value of the Monte Carlo test for the significance of $p(\tau \geq 0) = p(\tau > 0) + p(\tau = 0)$. }
\end{table}

\begin{table}
\begin{center}
{\footnotesize
\begin{tabular}{@{\extracolsep{0.7em}}llll}
\multicolumn{2}{c}{Prague} & \multicolumn{2}{c}{Stanford} \\ \cline{1-2}\cline{3-4}
Treebank & $p_0$ & Treebank & $p_0$ \\
\hline
{\bf Romanian} & 1 & {\bf Tamil} & 0.68\\
{\bf Japanese} & 0.97 & {\bf Telugu} & 0.62\\
{\bf Telugu} & 0.88 & {\bf Estonian} & 0.45\\
{\bf Tamil} & 0.84 & {\bf Bengali} & 0.29\\
{\bf Estonian} & 0.82 & {\bf Romanian} & 0.21\\
{\bf Arabic} & 0.34 & Turkish & 0.11\\
Italian & 0.32 & Greek(modern) & 0.11\\
Greek(modern) & 0.22 & Finnish & 0.091\\
{\bf Bengali} & 0.19 & Hindi & 0.086\\
Catalan & 0.16 & Italian & 0.051\\
Spanish & 0.14 & Bulgarian & 0.038\\
Finnish & 0.12 & Catalan & 0.035\\
Danish & 0.11 & Arabic & 0.033\\
Swedish & 0.085 & Basque & 0.03\\
Turkish & 0.07 & Spanish & 0.025\\
Russian & 0.061 & Latin & 0.022\\
Bulgarian & 0.059 & Danish & 0.018\\
Hindi & 0.052 & Hungarian & 0.016\\
English & 0.045 & Russian & 0.015\\
Hungarian & 0.032 & English & 0.015\\
Basque & 0.03 & Portuguese & 0.013\\
Portuguese & 0.025 & Czech & 0\\
Slovene & 0.022 & Dutch & 0\\
{\bf Persian} & 0.014 & German & 0\\
Czech & 0 & Greek(ancient) & 0\\
Dutch & 0 & Japanese & 0\\
German & 0 & Persian & 0\\
Greek(ancient) & 0 & Slovak & 0\\
Latin & 0 & Slovene & 0\\
Slovak & 0 & Swedish & 0\\
\end{tabular}
}
\end{center}
\caption{\label{sentence_lengths_without_crossings_table} $p_0$, the proportion of sentence lengths where $C=0$ for all sentences. Treebanks are sorted decreasingly by $p_0$. The treebanks where the null hypothesis could not be rejected according to Table \ref{dependencies_table} appear in boldface. }
\end{table}

\begin{table}
\begin{center}
{\footnotesize
\begin{tabular}{@{\extracolsep{0.7em}}lllllll}
 & \multicolumn{3}{c}{Prague} & \multicolumn{3}{c}{Stanford} \\ \cline{2-4}\cline{5-7}  	
 & mean & left p-value & right p-value & mean & left p-value & right p-value \\
\hline
$p_0$ & 0.8 & 1 & $10^{-6}$ & 0.4 & 1 & $8 \times 10^{-6}$ \\
$S$ & 2128.7 & $10^{-3}$ & 1 & 1288 & $6.5 \times 10^{-5}$ & 1\\
$M$ & 37.7 & 0.016 & 0.98 & 23.2 & $3.4 \times 10^{-5}$ & 1 \\
$\left< n \right>$ & 11.9 & 0.038 & 0.96 & 8.6 & $1.6 \times 10^{-4}$ & 1 \\
\end{tabular}
}
\end{center}
\caption{\label{meta_analysis_table} A meta-analysis of the subset of treebanks where $p(\tau \geq 0)$ is not significantly high with the help of one-sided Fisher randomization tests on the mean of a given treebank feature over that subset \cite{Conover1999a}. Four features are considered: $p_0$ (the proportion of sentence lengths where all sentences are planar), $S$ (the number of sentences), $M$ (the number of different sentence lengths) and $\left<n \right>$ (the mean length of sentences). p-values were estimated with the help of a Monte Carlo procedure over $10^6$ replicas and then rounded to leave only two significant digits. Means were rounded to leave only one decimal.}
\end{table}

\section{Discussion}

We have rejected the traditional hypothesis of crossings as being constrained independently from the dependency lengths in a large majority of treebanks (47 out of 60) thanks to a positive correlation between crossings ($C$) and dependency lengths ($D$) that holds across sentence lengths. The fact that the number of rejections depends on the annotation style (eight treebanks for Prague dependencies, five treebanks for Stanford dependencies) suggests that annotation criteria are crucial. Indeed, we have seen that there is a strong tendency for 
$C = 0$ across sentence lengths in those treebanks (Table \ref{sentence_lengths_without_crossings_table}).  
  
Before concluding prematurely that the minority of languages where the traditional view could not be rejected constitute evidence of an autonomous ban of crossings, some words of caution are necessary. First, we should reflect on the influence that syntactic dependency annotation criteria have had on the results due to:
\begin{itemize}
\item
A belief in a ban of crossings \cite{Sassano2005,Iwatate2008} or a principle of minimization of crossings. 
\item
Automatic conversions from phrase structure grammar to dependency treebanks \cite{et,ja}, where crossings could be less likely with respect to direct annotations based on dependency grammar.
\item
Annotation by automatic parsing followed by manual revision \cite{ta}, which can be biased due to either the parser not supporting crossings, or just having low recall for crossing dependencies, a common limitation even in modern non-projective parsers \cite{BjorkelundKuhn2012,GomNivCL2013}.
\item
The need of avoiding crossings to facilitate parsing by computers, as treebanks and annotation guidelines are often developed with this goal in mind \cite{ro,Begum2008}. 
\item
Cognitive considerations: dependency structures with fewer crossings being easier to understand by humans \cite{Purchase1997a,Huang2006a}.
\item
Aesthetical considerations: dependency structures with crossings being considered nicer than structures with crossings (see \cite{Kobourov2014a} and references therein). These preferences are supported by the cognitive considerations above. 
\end{itemize}
Second, we should also reflect on statistical factors:
\begin{itemize}
\item
The limited capacity of $D$ to predict crossings discussed above (Section \ref{modelling_section}).
\item
Insufficient sampling: the number of sentences ($S$) and the number of different sentence lengths ($M$) is significantly small in the subset of the treebanks where $p(\tau \geq 0)$ is not significantly high (Table \ref{meta_analysis_table}).
\item
A low mean sentence length. The point is that the chances for crossings are {\em a priori} lower in smaller sentences for various reasons. On the one hand, the size of the set of edges that may potentially cross grows with sentence length in general (Eq. \ref{cardinality_of_Q_equation} in the Appendix). On the other hand, the combination of three facts, i.e. 
\begin{itemize}
\item
The well-known tendency of $D$ to decrease as sentence length decreases \cite{Ferrer2004b,Ferrer2013c,Futrell2015a,Jiang2015a}
\item 
True values of $D$ are below chance \cite{Ferrer2004b,Ferrer2013c,Futrell2015a,Jiang2015a}
\item
The reduction of the probability that two dependencies cross by chance as they shorten (provided that they are sufficiently short) \cite{Ferrer2014c,Ferrer2014f}
\end{itemize} 
suggests that the abundance of $C = 0$ in some treebanks could be a side effect of the principle of dependency length minimization \cite{Ferrer2013e}, rather than an external imposition.
This possibility is supported by the significantly low mean sentence length that is found in the subset of treebanks where $p(\tau \geq 0)$ is not significantly high (Table \ref{meta_analysis_table}). However, this issue should be the subject of future research because dependency length minimization could be beaten by other word order principles at short sentence lengths \cite{Ferrer2014a}.  
\end{itemize} 
Halfway between annotation and statistical factors we find the decision of some treebanks' annotators to break complex sentences into simple clauses \cite{ro}. This procedure removes long distance dependencies, reduces mean sentence length and for the reasons reviewed above, could reduce the chance of crossings. 
By having examined a series of statistical caveats, we do not mean that they are the ultimate reason for the failure to reject the null hypothesis in some languages. Those factors, e.g., mean sentence length,  could be influenced by aspects such as modality (oral versus written) \cite{Gibson1966} or the genre of the sources used for the treebanks \cite{Kelih2006}. However, controlling for these aspects is beyond the scope of this article. For these reasons, it is convenient to be conservative and interpret the failure to reject the null model as a treebank-specific result that cannot be ascribed to a general property of the involved languages or an absence of dependency length minimization in them.  

Given all the preceding considerations,
our results and previous work \cite{Ferrer2014c,Ferrer2014f} provide support for the hypothesis that dependency crossings are a side effect of dependency lengths. By not requiring a belief in an autonomous ban of crossings \cite{Temperley2008a, Park2009a, Gildea2010a, Futrell2015a}, this hypothesis promises to help develop a more parsimonious theory of syntax. 



\section*{Appendix}

The traditional view could be recast as a simple model that predicts, given a sentence, a zero number of crossings. This deterministic model with no parameter could be generalized as a stochastic model with one parameter $a$ that defines the expected number of crossings. 
Suppose that $E[C|\mathit{sentence}]$ is the expectation of $C$ over all possible orderings of a sentence \cite{Ferrer2014f}. Then the traditional view could be defined as 
\begin{equation}
E[C | \mathit{sentence}] = a,
\label{naive_ban_of_crossings_model_equation}
\end{equation} 
where $a$ is a constant such that $a \geq 0$. $a = 0$ implies a ban of crossings because $C \geq 0$.
The parameter $a$ allows one to model crossings in languages with varying frequencies of crossings (from languages where there are no crossings to languages where crossings occur with a certain frequency). 

If the relevant information of a sentence is $D$, the sum of dependency lengths (see Fig. \ref{non_crossing_dependencies_figure} for examples of $D$), the alternative hypothesis could be modeled simply as \cite{Ferrer2014f}
\begin{equation}
E[C|D] = g(D),
\label{naive_simple_alternative_model_equation}
\end{equation}
where $g$ is a function of $D$,
and then the traditional hypothesis could be written as 
\begin{equation}
E[C|D] = a.
\end{equation}
A limitation of $E[C|D]$ is that it is defined over a set of possible linearizations that includes some that are very unlikely, cognitively harder or ``ungrammatical''. In this article, we focus on real linearizations and therefore we consider $E_{TB}[C|D]$, the expectation of $C$ given $D$ over the ensemble of linearizations of the sentences of a treebank (TB). $E_{TB}[C|D]$ needs to be refined: the distribution of $D$ depends on the length of the sentence and then values of $D$ from sentences of different length should not be mixed \cite{Ferrer2013c}. The same kind of problem is also likely to concern $C$. 
For this reason, instead of  
$E_{TB}[C|D]$, we choose $E_{TB}[C|n, D]$, i.e. the expectation of $C$ conditioning on sentences of the treebank that have length $n$ and their sum of dependency lengths is $D$.

Now we will explain why $a_{TB}$ depends on $n$ by means of a key concept of crossing theory: $Q$, namely the set of pairs of edges that may potentially cross when their vertices are arranged linearly \cite{Ferrer2013b,Gomez2016a}. By definition, $C \leq |Q|$, the cardinality of $Q$.
When $n \geq 1$, we have that \cite{Ferrer2013b},
\begin{equation}
|Q| \leq \frac{n}{2}\left( n - 1 - \left< k^2 \right> \right),
\end{equation}
where $\left< k^2 \right>$ is the degree's second moment about zero. 
Knowing that $\left< k^2 \right> \leq \left< k^2 \right>^{linear}$, the value of $\left< k^2 \right>$ of a linear tree, and that $\left< k^2 \right>^{linear} = 4 - 6/n$ (when $n \geq 2$) \cite{Ferrer2013b}, we finally obtain 
\begin{equation}
C \leq |Q| \leq \frac{n}{2}(n-5) + 3
\label{cardinality_of_Q_equation}
\end{equation} 
for $n\geq 2$.
For instance, this implies that $a_{TB}(n) = 0$ for $n < 4$ (since $C=0$ in this case \cite{Ferrer2013b}) and that $0 \leq a_{TB}(4) \leq 1$. It is clear that one cannot set $a_{TB}(n)$ to a number greater than $2$ when $n \leq 4$ because it cannot be reached by $E_{TB}[C | n, D]$. In general ($n\geq 2$), $a_{TB}(n) > \frac{n}{2}(n-5) + 3$ is impossible to achieve. This is why $a_{TB}$ depends on $n$ {\em a priori}.

\section*{Acknowledgments}

We thank two anonymous reviewers for their valuable comments.  
We are also grateful to R. Levy for helpful comments and discussions.
RFC is funded by the grants 2014SGR 890 (MACDA) from AGAUR (Generalitat de Catalunya) and also
the APCOM project (TIN2014-57226-P) from MINECO (Ministerio de Economia y Competitividad).
CGR is partially funded by the TELEPARES-UDC project (FFI2014-51978-C2-2-R) from MINECO and an Oportunius program grant from Xunta de Galicia.

\bibliographystyle{naturemag_noURL}

\bibliography{biblio}

\end{document}